\documentclass{article}
\PassOptionsToPackage{authoryear, round, compress}{natbib}
\usepackage[final]{nips_2017} 

\usepackage[dvipsnames]{xcolor}

\usepackage[utf8]{inputenc} 
\usepackage[T1]{fontenc}    
\usepackage{hyperref}       
\usepackage{url}            
\usepackage{booktabs}       
\usepackage{amsfonts}       
\usepackage{nicefrac}       
\usepackage{microtype}      
\usepackage{amsmath,amssymb,amsthm,mathrsfs,amsfonts,dsfont}
\usepackage{graphicx}
\usepackage{wrapfig}

\usepackage{lscape}

\newcommand{\R}[0]{\mathds{R}} 
\renewcommand{\vec}[1]{\mathbf{#1}} 
\newcommand{\mat}[1]{\mathbf{#1}} 
\newcommand{\inv}[0]{^{-1}}
\newcommand{\T}[0]{^{\top}}
\newcommand{\KL}[0]{\textsc{KL}}

\newcommand{\sz}[0]{z} 
\newcommand{\modeling}[0]{modelling} 

\newcommand{\optimization}[0]{optimi\sz ation} 
\newcommand{\optimize}[0]{optimi\sz e} 
\newcommand{\optimized}[0]{optimi\sz ed}
\newcommand{\optimizing}[0]{optimi\sz ing} 

\newcommand{\generalizations}[0]{generali\sz ations}

\newcommand{\generalized}[0]{generali\sz ed} 

\newcommand{\parameterized}[0]{parameteri\sz ed} 
\newcommand{\parameterize}[0]{parameteri\sz e} 
\newcommand{\reparameterization}[0]{re-parameteri\sz ation }

\newcommand{\minimizing}[0]{minimi\sz ing} 
\newcommand{\maximize}[0]{maximi\sz e}

\newcommand{\characterized}[0]{characteri\sz ed}

\newcommand{\initialization}[0]{initiali\sz ation} 
\newcommand{\initialize}[0]{initiali\sz e} 

\newcommand{\marginalize}[0]{marginali\sz e} 
\newcommand{\marginalization}[0]{marginali\sz ation} 

\newcommand{\emphasize}[0]{emphasi\sz e}

\newcommand{\factorizes}[0]{factori\sz es} 
\newcommand{\factorized}[0]{factori\sz ed} 
\newcommand{\factorization}[0]{factori\sz ation} 

\newcommand{\amortize}[0]{amorti\sz e}

\newcommand{\layer}[2]{#1 ^{{#2}}} 
\newcommand{\kxx}[2]{k({#1},{#2})} 

\newtheorem{remark}{Remark}

\newcommand{\red}[1]{\color{red} #1 \color{black}}
\newcommand{\blue}[1]{\color{blue} #1 \color{black}}
\newcommand{\orange}[1]{\color{orange} #1 \color{black}}
\newcommand{\green}[1]{\color{OliveGreen} #1 \color{black}}

\title{Doubly Stochastic Variational Inference \\for Deep Gaussian Processes}

\author{
  Hugh Salimbeni \\
  Imperial College London and PROWLER.io\\
  \texttt{hrs13@ic.ac.uk} \\
  \And
  Marc Peter Deisenroth \\
  Imperial College London and PROWLER.io\\
  \texttt{m.deisenroth@imperial.ac.uk} \\
 }

\begin{document}
    
\maketitle

\vspace{-0.3cm}

\begin{abstract}


Gaussian processes (GPs) are a good choice for function approximation as they are flexible, robust to overfitting, and provide well-calibrated predictive uncertainty.
%
%
Deep Gaussian processes (DGPs) are multi-layer \generalizations~of GPs, but inference in these models has proved challenging.
%
%
%
Existing approaches to inference in DGP models assume approximate posteriors that force independence between the layers, and do not work well in practice.
%
We present a doubly stochastic variational inference algorithm that does not force independence between layers.
%
With our method of inference we demonstrate that a DGP model can be used effectively on data ranging in size from hundreds to a billion points. We provide strong empirical evidence that our inference scheme for DGPs works well in practice in both classification and regression.

\end{abstract}

\section{Introduction}
Gaussian processes (GPs) achieve state-of-the-art performance in a range of applications including robotics~\citep{Ko2009,Deisenroth2011c}, geostatistics~\citep{diggle2007springer}, numerics~\citep{Briol2015ProbabilisticAnalysis}, active sensing~\citep{Guestrin2005Near-optimalProcesses} and \optimization~\citep{Snoek2012}. 
A Gaussian process is defined by its mean and covariance function. 
In some situations prior knowledge can be readily incorporated into these functions.
Examples include periodicities in climate \modeling~\citep{Rasmussen2006}, change-points in time series data~\citep{Garnett2009SequentialChangepoints} and simulator priors for robotics~\citep{Cutler2015EfficientPriors}. 
%
In other settings, GPs are used successfully as black-box function approximators.
There are compelling reasons to use GPs, even when little is known about the data: a GP  grows in complexity to suit the data; a GP is robust to overfitting while providing reasonable error bars on predictions; a GP can model a rich class of functions with few  hyperparameters. 

Single-layer GP models are limited by the expressiveness of the kernel/covariance function. 
To some extent kernels can be learned from data, but inference over a large and richly \parameterized~space of kernels is expensive, and approximate methods may be at risk of overfitting. 
Optimization of the marginal likelihood with respect to hyperparameters approximates Bayesian inference only if the number of hyperparameters is small~\citep{Mackay1999ComparisonHyperparameters}. 
Attempts to use, for example, a highly \parameterized~neural network as a kernel function~\citep{Calandra2016ManifoldRegression,Wilson2016DeepLearning} incur the downsides of deep learning, such as the need for application-specific architectures and regularization techniques. 
Kernels can be combined through sums and products~\citep{Duvenaud2013StructureSearch} to create more expressive compositional kernels, but this approach is limited to simple base kernels, and their \optimization~is expensive.

A Deep Gaussian Process (DGP) is a hierarchical composition of GPs that can overcome the limitations of standard (single-layer) GPs while retaining the advantages.
DGPs are richer models than standard GPs, just as deep networks are richer than \generalized~linear models. 
In contrast to models with highly \parameterized~kernels, DGPs learn a representation hierarchy non-parametrically with very few hyperparmeters to \optimize. 

Unlike their single-layer counterparts, DGPs have proved difficult to train. 
The mean-field variational approaches used in previous work~\citep{Damianou2013DeepProcesses,Mattos2016RecurrentProcesses,Dai2016VariationalProcesses} make strong independence and Gaussianity assumptions. 
The true posterior is likely to exhibit high correlations between layers, but mean-field variational approaches are known to severely underestimate the variance in these situations~\citep{Turner2011TwoModels}.

In this paper, we present a variational algorithm for inference in DGP models that does \emph{not} force independence or Gaussianity between the layers.
In common with many state-of-the-art GP approximation schemes we start from a sparse inducing point variational framework~\citep{Matthews2016OnProcesses} to achieve computational tractability \emph{within} each layer, but we do not force independence \emph{between} the layers. 
Instead, we use the exact model conditioned on the inducing points as a variational posterior. This posterior has the same structure as the full model, and in particular it maintains the correlations between layers. 
Since we preserve the non-linearity of the full model in our variational posterior we lose analytic tractability. We overcome this difficulty by sampling from the variational posterior, introducing the first source of stochasticity. This is computationally straightforward due to an important property of the sparse variational posterior marginals: the marginals conditioned on the layer below depend only on the corresponding inputs. It follows that samples from the marginals at the top layer can be obtained without computing the full covariance \emph{within} the layers.
%
%
We are primarily interested in large data applications, so we further subsample the data in minibatches. This second source of stochasticity allows us to scale to arbitrarily large data.

We demonstrate through extensive experiments that our approach works well in practice.
We provide results on benchmark regression and classification data problems, and also demonstrate the first DGP application to a dataset with a billion points.
Our experiments confirm that DGP models are never worse than single-layer GPs, and in many cases significantly better. Crucially, we show that  additional layers do not incur overfitting, even with small data.


\section{Background}
In this section, we present necessary background on single-layer Gaussian processes and sparse variational inference, followed by the definition of the deep Gaussian process model. Throughout we \emphasize~a particular property of sparse approximations: the sparse variational posterior is itself a Gaussian process, so the marginals depend only on the corresponding inputs.


\subsection{Single-layer Gaussian Processes}
We consider the task of inferring a stochastic function $f: \R^{D} \to \R$, given a likelihood $p(y|f)$ and a set of $N$ observations $\vec y = (y_1, \dots, y_N)\T$ at design locations $\vec X = (\vec x_1, \dotsc, \vec x_N)\T$. We place a GP prior on the function $f$ that models all function values as jointly Gaussian, with a covariance function $k: \R^{D} \times \R^{D} \to \R$ and a mean function $m: \R^{D} \to \R$. 
We further define an additional set of $M$ inducing locations $\vec Z = (\vec z_1, \dots, \vec z_M)\T$. We use the notation $\vec f = f(\vec X)$ and $\vec u =f(\vec Z)$ for the function values at the design and inducing points, respectively. We define also $[m (\vec X)]_i = m(\vec x_i)$ and $[k(\vec X, \vec Z)]_{ij} = k(\vec x_i, \vec z_j)$. By the definition of a GP, the joint density $p(\vec f, \vec u)$ is a Gaussian whose mean is given by the mean function evaluated at every input $(\vec X, \vec Z)\T$, and the corresponding covariance is given by the covariance function evaluated at every pair of inputs. The joint density of $\vec y,\vec f$ and $\vec u$ is  
\begin{align}
\label{eq:GP_density}
p(\vec y, \vec f, \vec u) =
\underbrace{
p(\vec f | \vec u; \vec X, \vec Z)
p(\vec u ; \vec Z)
}_\text{GP prior} 
\underbrace{
\prod\nolimits_{i=1}^{N} p(y_i|f_i)
}_\text{likelihood}\,.
\end{align}
In~\eqref{eq:GP_density} we \factorized~the joint GP prior $p(\vec f, \vec u; \vec X, \vec Z)$ \footnote{Throughout this paper we use the semi-colon notation to clarify the input locations of the corresponding function values, which will become important later when we discuss multi-layer GP models. For example, $p(\vec f | \vec u; \vec X, \vec Z)$ indicates that the input locations for $\vec f$ and $\vec u$ are $\vec X$ and $\vec Z$, respectively.} into the prior $p(\vec u) = \mathcal{N}(\vec u |  m(\vec Z), \kxx{\vec Z}{\vec Z})$ and the conditional $p(\vec f | \vec u; \vec X, \vec Z) = \mathcal{N}(\vec f | \boldsymbol \mu, \boldsymbol \Sigma)$, where for $i,j=1,\dotsc,N$
\begin{align}
\label{eq:mu_conditional}
[\boldsymbol\mu]_i & =  m(\vec x_i) + \boldsymbol \alpha(\vec x_i)\T (\vec u - m(\vec Z))\,, \\
\label{eq:Sigma_conditional}
[\boldsymbol\Sigma]_{ij} &=  k(\vec x_i, \vec x_j) - \boldsymbol \alpha(\vec x_i)\T \kxx{\vec Z}{ \vec Z}\boldsymbol \alpha(\vec x_j)\,,
\end{align}
with $\boldsymbol \alpha(\vec x_i) =\kxx{\vec Z}{\vec Z}\inv \kxx{\vec Z}{\vec x_i}$. Note that the conditional mean $\boldsymbol\mu$ and covariance $\boldsymbol\Sigma$ defined via~\eqref{eq:mu_conditional} and~\eqref{eq:Sigma_conditional}, respectively, take the form of mean and covariance functions of the inputs $\vec x_i$.
%
%
Inference in the model~\eqref{eq:GP_density} is possible in closed form when the likelihood $p(y|f)$ is Gaussian, but the computation scales cubically with $N$. 

We are interested in large datasets with non-Gaussian likelihoods. Therefore, we seek a variational posterior to overcome both these difficulties simultaneously. Variational inference seeks an approximate posterior $q(\vec f, \vec u)$ by \minimizing~the Kullback-Leibler divergence $\KL[q||p]$ between the variational posterior $q$ and the true posterior $p$. Equivalently, we \maximize~the lower bound on the marginal likelihood (evidence)
\begin{align}
\mathcal{L}=\mathbb{E}_{q(\vec f, \vec u)}\left[\log \frac{p(\vec y, \vec f, \vec u)}{q(\vec f, \vec u)}\right]\,,
\label{eq:ELBO}
\end{align}
where $p(\vec y, \vec f, \vec u)$ is given in~\eqref{eq:GP_density}. We follow~\cite{Hensman2013GaussianData} and choose a variational posterior 
\begin{align}
\label{eq:single_layer_variational_posterior}
q(\vec f, \vec u)=p(\vec f | \vec u; \vec X, \vec Z)q(\vec u)\,,
\end{align}
where $q(\vec u)=\mathcal{N}(\vec u| \vec m, \vec S)$.  Since both terms in the variational posterior are Gaussian, we can analytically \marginalize~$\vec u$, which yields
\begin{align}
\label{eq:q_f_full}
q(\vec f | \vec m, \vec S ; \vec X, \vec Z) &= \int p(\vec f | \vec u; \vec X, \vec Z) q(\vec u) d\vec u 
= \mathcal{N}(\vec f|  
\tilde{\boldsymbol \mu},
\tilde{\boldsymbol \Sigma})\,.
\end{align}
Similar to~\eqref{eq:mu_conditional} and~\eqref{eq:Sigma_conditional}, the expressions for $\tilde{\boldsymbol \mu}$ and $\tilde{\boldsymbol \Sigma}$ can be written as mean and covariance functions of the inputs. To \emphasize~this point we define
\begin{align}
\mu_{\vec m, \mat Z}(\vec x_i) &= m(\vec x_i) + \boldsymbol\alpha(\vec x_i)\T (\vec m -  m(\vec Z))\,, 
\label{eq:definition mu augmented}\\
\Sigma_{\mat S, \mat Z}(\vec x_i, \vec x_j) &=  k(\vec x_i, \vec x_j) - \boldsymbol\alpha(\vec x_i)\T (\kxx{\vec Z}{\vec Z} - \mat S) \boldsymbol\alpha(\vec x_j)\,.
\label{eq:definition Sigma augmented}
\end{align}
With these functions we define $[\tilde{\boldsymbol \mu}]_i = \mu_{\vec m, \mat Z}(\vec x_i)$ and $[\tilde{\boldsymbol \Sigma}]_{ij} = \Sigma_{\mat S, \mat Z}(\vec x_i, \vec x_j)$. We have written the mean and covariance in this way to make the following observation clear.


\begin{remark}
The $f_i$ marginals of the variational posterior~\eqref{eq:q_f_full} depend only on the corresponding inputs $\vec x_i$. Therefore, we can write the $i$th marginal of $q(\vec f | \vec m, \vec S; \vec X, \vec Z)$ as 
\begin{align}
\label{eq:q_marginal}
q(f_i | \vec m, \vec S; \vec X, \vec Z) = q(f_i | \vec m, \vec S; \vec x_i, \vec Z) = \mathcal{N}(f_i | \mu_{\vec m, \mat Z}(\vec x_i), \Sigma_{\vec S, \mat Z}(\vec x_i, \vec x_i))\,.
\end{align}
\end{remark}
Using our variational posterior~\eqref{eq:single_layer_variational_posterior} the lower bound~\eqref{eq:ELBO} simplifies considerably since (a) the conditionals $p(\vec f|\vec u; \vec X, \vec Z)$ inside the logarithm cancel and (b) the likelihood expectation requires only the variational marginals. We obtain
\begin{align}
\label{eq:lower_bound_basic}
\mathcal{L} = 
\sum\nolimits_{i=1}^{N} 
\mathbb{E}_{q(f_i | \vec m, \vec S; \vec x_i, \vec Z)}
[\log p(y_i|f_i)] 
- \KL[q(\vec u)||p(\vec u)]\,.
\end{align}

The final (univariate) expectation of the log-likelihood can be computed analytically in some cases, with quadrature~\citep{Hensman2015MCMCProcesses} or through Monte Carlo sampling~\citep{Bonilla2016GenericModels,Gal2015LatentData}. Since the bound is a sum over the data, an unbiased estimator can be obtained through minibatch subsampling. This permits inference on large datasets. In this work we refer to a GP with this method of inference as a \emph{sparse GP (SGP)}. 

The variational parameters ($\vec Z$, $\vec m$ and $\vec S$) are found by maximizing the lower bound~\eqref{eq:lower_bound_basic}. This maximization is guaranteed to converge since $\mathcal{L}$ is a lower bound to the marginal likelihood $p(\vec y|\vec X)$. We can also learn model parameters (hyperparameters of the kernel or likelihood) through the maximization of this bound, though we should exercise caution as this introduces bias because the bound is not uniformly tight for all settings of hyperparameters~\citep{Turner2011TwoModels}

So far we have considered scalar outputs $y_i\in\R$. In the case of $D$-dimensional outputs $\vec y_i\in\R^D$ we define $\vec Y$ as the matrix with $i$th row containing the $i$th observation $\vec y_i$. Similarly, we define  $\vec F$ and $\vec U$. If each output is an independent GP we have the GP prior $\prod_{d=1}^{D}p(\vec F_d | \vec U_d; \vec X, \vec Z)p(\vec U_d; \vec Z)$, which we abbreviate as $p(\vec F | \vec U; \vec X, \vec Z)p(\vec U; \vec Z)$ to lighten the notation.

\subsection{Deep Gaussian Processes} 
A DGP~\citep{Damianou2013DeepProcesses} defines a prior recursively on vector-valued stochastic functions $\layer{F}{1}, \dots, \layer{F}{L}$. The prior on each function $\layer{F}{l}$ is an independent GP in each dimension, with input locations given by the noisy corruptions of the function values at the next layer: the outputs of the GPs at layer $l$ are $F^{l}_d$, and the corresponding inputs are $F^{l-1}$. The noise between layers is assumed i.i.d. Gaussian. Most presentations of DGPs ~\citep[see, e.g.][]{Damianou2013DeepProcesses, Bui2016DeepPropagation} explicitly \parameterize~the noisy corruptions separately from the outputs of each GP. Our method of inference does not require us to \parameterize~these variables separately. For notational convenience, we therefore absorb the noise into the kernel $k_{noisy}(\vec x_i, \vec x_j) = k(\vec x_i, \vec x_j) + \sigma_l^2 \delta_{ij}$, where $\delta_{ij}$ is the Kronecker delta, and $\sigma_l^2$ is the noise variance between layers.
We use $\layer{D}{l}$ for the dimension of the outputs at layer $l$. As with the single-layer case, we have inducing locations $\vec Z^{l-1}$ at each layer and inducing function values $\vec U^l$ for each dimension. 

An instantiation of the process has the joint density
\begin{align}
\label{eq:noisy_DGP_def}
p(
\vec Y,
\{
\layer{\vec F}{l}, 
\layer{\vec U}{l}
\}_{l=1}^{L}) =
\underbrace{
\prod\nolimits_{i=1}^{N}p(\vec y_i|\layer{\vec f_i}{L})
}_{\text{likelihood}}
\underbrace{
\prod\nolimits_{l=1}^{L} 
p( \layer{\vec{F}}{l}|  \layer{\vec {U}}{l};  \layer{\vec{F}}{l-1},  \layer{\vec{Z}}{l-1})
p( \layer{\vec{U}}{l} ;  \layer{\vec{Z}}{l-1})
}_{\text{DGP prior}}\,,
\end{align}
where we define ${\layer{\vec F}{0}} = \vec X$. Inference in this model is intractable, so approximations must be used.

The original DGP presentation~\citep{Damianou2013DeepProcesses} uses a variational posterior that maintains the exact model conditioned on $\vec U^l$, but further forces the inputs to each layer to be independent from the outputs of the previous layer. The noisy corruptions are \parameterized~separately, and the variational distribution over these variables is a fully \factorized~Gaussian. This approach requires $2N(D^1+\dots + D^{L-1})$ variational parameters but admits a tractable lower bound on the log marginal likelihood if the kernel is of a particular form. A further problem of this bound is that the density over the outputs is simply a single layer GP with independent Gaussian inputs. Since the posterior loses all the correlations between layers it cannot express the complexity of the full model and so is likely to underestimate the variance. In practice, we found that \optimizing~the objective in~\cite{Damianou2013DeepProcesses} results in layers being `turned off' (the signal to noise ratio tends to zero). 
In contrast, our posterior retains the full conditional structure of the true model. We sacrifice analytical tractability, but due to the sparse posterior \emph{within} each layer we can sample the bound using univariate Gaussians.

\section{Doubly Stochastic Variational Inference}
In this section, we propose a novel variational posterior and demonstrate a method to obtain unbiased samples from the resulting lower bound. 
The difficulty with inferring the DGP model is that there are complex correlations both \emph{within} and \emph{between} layers. Our approach is straightforward: we use sparse variational inference to simplify the correlations \emph{within} layers, but we maintain the correlations \emph{between} layers. The resulting variational lower bound cannot be evaluated analytically, but we can draw unbiased samples efficiently using univariate Gaussians. We \optimize~our bound stochastically.

We propose a posterior with three properties. Firstly, the posterior maintains the exact model, conditioned on $\vec U^l$. Secondly, we assume that the posterior distribution of $\{\vec U^l\}_{l=1}^{L}$ is \factorized~between layers (and dimension, but we suppress this from the notation).
Therefore, our posterior takes the simple \factorized~form
\begin{align}
\label{eq:our_posterior}
q(
\{
\layer{\vec F}{l}, 
\layer{\vec U}{l}
\}_{l=1}^{L}) = 
\prod\nolimits_{l=1}^{L} 
p( \layer{\vec{F}}{l}|  \layer{\vec {U}}{l};  \layer{\vec{F}}{l-1},  \layer{\vec{Z}}{l-1})
q( \layer{\vec{U}}{l})\,.
\end{align}
Thirdly, and to complete specification of the posterior, we take $q(\layer{\vec{U}}{l})$ to be a Gaussian 
with mean $\vec m^l$ and variance $\vec S^l$. A similar posterior was used in \cite{Hensman2014NestedProcesses} and \cite{Dai2016VariationalProcesses}, but each of these works contained additional terms for the noisy corruptions at each layer. 

%
%
%
%
As in the single layer SGP, we can \marginalize~the inducing variables from each layer analytically. After this \marginalization~we obtain following distribution, which is fully coupled within and between layers:
\begin{align}
\label{eq:u_marginal}
q(
\{
\layer{\vec F}{l}
\}_{l=1}^{L}) = 
\prod\nolimits_{l=1}^{L} 
q(\layer{\vec{F}}{l}| \layer{\vec m}{l}, \layer{\vec S}{l}; \layer{\vec F}{l-1}, \layer{\vec Z}{l-1})  =\prod\nolimits_{l=1}^{L}  \mathcal N(\vec F^l|\layer{\tilde{\boldsymbol\mu}}{l},\layer{\tilde{\boldsymbol\Sigma}}{l})\,.
\end{align}
Here, 
$q(\layer{\vec{F}}{l}| \layer{\vec m}{l}, \layer{\vec S}{l}; \layer{\vec F}{l-1}, \layer{\vec Z}{l-1})$ 
is as in \eqref{eq:q_f_full}. 
Specifically, it is a Gaussian with mean and variance 
$\layer{\tilde{\boldsymbol \mu}}{l}$
and $\layer{\tilde{\boldsymbol \Sigma}}{l}$, 
where 
$[\layer{\tilde{\boldsymbol \mu}}{l}]_i = \mu_{\layer{\vec m}{l}, \layer{\vec Z}{l-1}} (\layer{\vec f_i}{l})$ 
and $[\layer{\tilde{\boldsymbol \Sigma}}{l}]_{ij} = \Sigma_{\layer{\vec S}{l}, \layer{\vec Z}{l-1}} (\layer{\vec f_i}{l}, \layer{\vec f_j}{l})$ 
(recall that $\layer{\vec f_i}{l}$ is the $i$th row of $\layer{\vec F}{l}$).
Since~\eqref{eq:our_posterior} is a product of terms that each take the form of the SGP variational posterior~\eqref{eq:single_layer_variational_posterior}, we have again the property that within each layer the marginals depend on only the corresponding inputs. In particular, $\layer{\vec f_i}{L}$ depends only on $\layer{\vec f_i}{L-1}$, which in turn depends only on $\layer{\vec f_i}{L-2}$, and so on. Therefore, we have the following property:
\begin{remark}
The $i$th marginal of the final layer of the variational DGP posterior \eqref{eq:our_posterior} depends only on the $i$th marginals of all the other layers. That is, 
\begin{align}
\label{eq:marginals_DGP_posterior}
q(\vec f_i^L) = \int \prod\nolimits_{l=1}^{L-1}  q(\layer{\vec f_i}{l}|\layer{\vec m}{l}, \layer{\vec S}{l}; \layer{\vec f_i}{l-1}, \layer{\vec Z}{l-1}) d\layer{\vec f_i}{l}\,.
\end{align}
\end{remark}

The consequence of this property is that taking a sample from $q(\layer{\vec f_i}{L})$ is straightforward, and furthermore we can perform the sampling using only univariate unit Gaussians using the `\reparameterization~trick' \citep{Rezende2014StochasticModels,Kingma2015VariationalTrick}. Specifically, we first sample $\boldsymbol\epsilon_i^l \sim \mathcal{N}(\mathbf 0, \mathbf I_{D^l})$ and then recursively draw the sampled variables $\layer{\vec{\hat{f}}_i}{l}\sim q(\layer{\vec f_i}{l}|\layer{\vec m}{l}, \layer{\vec S}{l}; \layer{\vec {\hat{f}}_i}{l-1}, \layer{\vec Z}{l-1})$ for $l=1,\dotsc ,L-1$ as
%
\begin{align}
\label{eq:sampling}
\layer{\vec{\hat{f}}_i}{l} = \mu_{\layer{\vec m}{l}, \layer{\vec Z}{l-1}} (\layer{\vec {\hat{f}}_i}{l-1}) +  \boldsymbol\epsilon_i^{l} \odot \sqrt{\Sigma_{\layer{\vec S}{l}, \layer{\vec Z}{l-1}} (\layer{\vec {\hat{f}}_i}{l-1}, \layer{\vec {\hat{f}}_i}{l-1})}\,,
\end{align}
where the terms in~\eqref{eq:sampling} are $D^l$-dimensional and the square root is element-wise. For the first layer we define $\layer{\vec {\hat{f}}_i}{0} := \vec x_i$. 

\paragraph{Efficient computation of the evidence lower bound} The evidence lower bound of the DGP is
\begin{align}
\label{eq:dgp lower bound general}
\mathcal{L}_{DGP} = \mathbb{E}_{q(\{\layer{\vec F}{l}, \layer{\vec U}{l}\}_{l=1}^L)}\left[\frac{p(\vec Y,\{\layer{\vec F}{l}, \layer{\vec U}{l}\}_{l=1}^L)}{q(\{\layer{\vec F}{l}, \layer{\vec U}{l}\}_{l=1}^L)}\right]\,.
\end{align}
Using~\eqref{eq:noisy_DGP_def} and \eqref{eq:our_posterior} for the corresponding expressions in~\eqref{eq:dgp lower bound general}, we obtain after some re-arranging
\begin{align}
\label{eq:lower_bound_ours}
\mathcal{L}_{DGP} = 
\sum_{i=1}\nolimits^{N}
\mathbb{E}_{q(\layer{\vec f_{i}}{L})}
[\log p(\vec y_n|\layer{\vec f_n}{L})]
-\sum\nolimits_{l=1}^{L} 
\KL[q(\layer{\vec U}{l}) || p(\layer{\vec U}{l}; \layer{\vec Z}{l-1})]\,,
\end{align}
where we exploited the exact marginalization of the inducing variables~\eqref{eq:u_marginal} and the property of the marginals of the final layer~\eqref{eq:marginals_DGP_posterior}. A detailed derivation is provided in the supplementary material. This bound has complexity  $\mathcal{O}(NM^2(D^1+\dots+D^L))$ to evaluate. 

We evaluate the bound~\eqref{eq:lower_bound_ours} approximately using two sources of stochasticity. Firstly, we approximate the expectation with a Monte Carlo sample from the variational posterior~\eqref{eq:marginals_DGP_posterior}, which we compute according to \eqref{eq:sampling}. Since we have \parameterized~this sampling procedure in terms of isotropic Gaussians, we can compute unbiased gradients of the bound~\eqref{eq:lower_bound_ours}. Secondly, since the bound \factorizes~over the data we achieve scalability through sub-sampling the data. Both stochastic approximations are \emph{unbiased}. 

\paragraph{Predictions}
To predict we sample from the variational posterior changing the input locations to the test location $\vec x_*$. We denote the function values at the test location as $\layer{\vec f_*}{l}$. To obtain the density over $\layer{\vec f_*}{L}$ we use the Gaussian mixture
\begin{align}
q(\layer{\vec f_*}{L}) \approx \frac{1}{S}\sum\nolimits_{s=1}^{S}  q(\layer{\vec f_*}{L}|\layer{\vec m}{L}, \layer{\vec S}{L}; \layer{{\vec f_*^{(s)}}}{L-1}, \layer{\vec Z}{L-1})\,,
\end{align}
where we draw $S$ samples $\layer{{\vec f_*^{(s)}}}{L-1}$ using~\eqref{eq:sampling}, but replacing the inputs $\vec x_i$ with the test location $\vec x_*$.

\paragraph{Further Model Details}
While GPs are often used with a zero mean function, we consider such a choice inappropriate for the inner layers of a DGP. Using a zero mean function causes difficulties with the DGP prior as each GP mapping is highly non-injective. This effect was analyzed in~\cite{Duvenaud2014AvoidingNetworks} where the authors suggest adding the original input $\vec X$ to each layer. 
Instead, we consider an alternative approach and include a linear mean function $m(\vec X)=\vec X\vec W$ for all the inner layers. If the input and output dimension are the same we use the identity matrix for $\vec W$, otherwise we compute the SVD of the data and use the top $D^{l}$ left eigenvectors sorted by singular value (i.e. the PCA mapping). With these choices it is effective to \initialize~all inducing mean values $\vec m^{l} = \vec 0$. This choice of mean function is partly inspired by the `skip layer' approach of the ResNet~\citep{He2016DeepRecognition} architecture.

\section{Results}
We evaluate our inference method on a number of benchmark regression and classification datasets. We stress that we are interested in models that can operate in both the small and large data regimes, with little or no hand tuning. All our experiments were run with exactly the same hyperparameters and initializations. See the supplementary material for details. We use $\text{min}(30, \layer{D}{0})$ for all the inner layers of our DGP models, where $\layer{D}{0}$ is the input dimension, and the RBF kernel for all layers. 


\paragraph{Regression Benchmarks} We compare our approach to other state-of-the-art methods on 8 standard small to medium-sized UCI benchmark datasets. Following common practice \cite[e.g.][]{Adams2015ProbabilisticNetworks} we use 20-fold cross validation with a 10\% randomly selected held out test set and scale the inputs and outputs to zero mean and unit standard deviation within the training set (we restore the output scaling for evaluation). While we could use any kernel, we choose the RBF kernel with a lengthscale for each dimension for direct comparison with~\cite{Bui2016DeepPropagation}. 
\begin{figure}[tb]
\includegraphics[width=\textwidth]{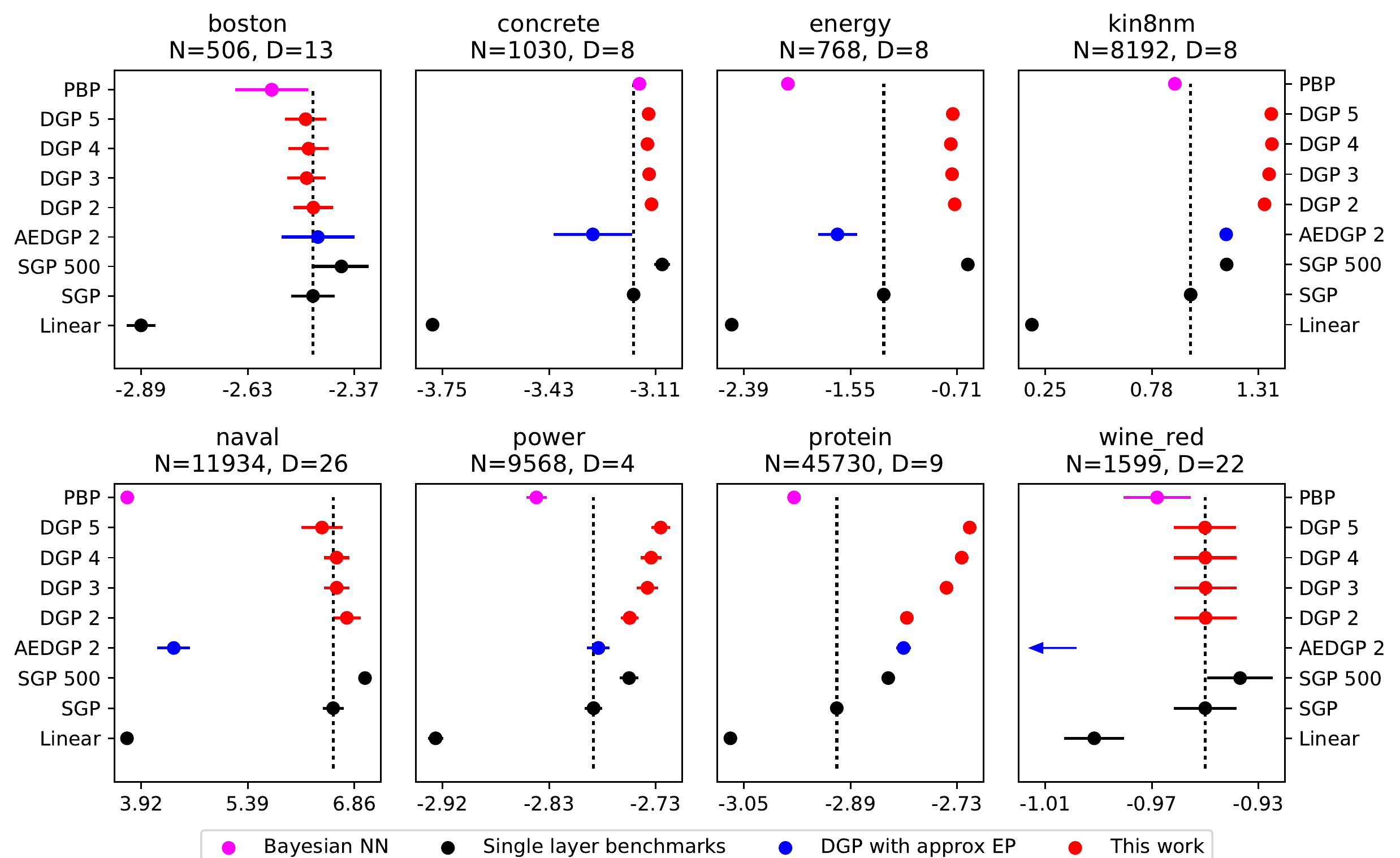}
\caption{Regression test log-likelihood results on benchmark datasets. Higher (to the right) is better. 
The sparse GP with the same number of inducing points is highlighted as a baseline.
}\label{fig:reg_results}
\end{figure}
The test log-likelihood results are shown in Fig.~\ref{fig:reg_results}. We compare our models of 2, 3, 4 and 5 layers (DGP 2--5), each with 100 inducing points, with (stochastically \optimized) sparse GPs~\citep{Hensman2013GaussianData} with 100 and 500 inducing points points (SGP, SGP 500). 
We compare also to a two-layer Bayesian neural network with ReLu activations, 50 hidden units (100 for protein and year), with inference by probabilistic backpropagation~\citep{Adams2015ProbabilisticNetworks} (PBP). The results are taken from~\cite{Adams2015ProbabilisticNetworks} and were found to be the most effective of several other methods for inferring Bayesian neural networks. We compare also with a DGP model with approximate expectation propagation (EP) for inference~\citep{Bui2016DeepPropagation}. Using the authors' code \footnote{\url{https://github.com/thangbui/deepGP_approxEP}} we ran a DGP model with 1 hidden layer using approximate expectation propagation~\citep{Bui2016DeepPropagation} (AEPDGP 2). We used the input dimension for the hidden layer for a fair comparison with our models\footnote{We note however that in~\cite{Bui2016DeepPropagation} the inner layers were 2D, so the results we obtained are not directly comparable to those reported in~\cite{Bui2016DeepPropagation}}. We found the time requirements to train a 3-layer model with this inference prohibitive. 
Plots for test RMSE and further results tables can be found in the supplementary material.

On five of the eight datasets, the deepest DGP model is the best. On `wine', `naval' and `boston' our DGP recovers the single-layer GP, which is not surprising: `boston' is very small, `wine' is near-linear (note the proximity of the linear model and the scale) and `naval' is \characterized~by extremely high test likelihoods (the RMSE on this dataset is less than 0.001 for all SGP and DGP models), i.e. it is a very `easy' dataset for a GP. The Bayesian network is not better than the sparse GP for any dataset and significantly worse for six. The Approximate EP inference for the DGP models is also not competitive with the sparse GP for many of the datasets, but this may be because the initializations were designed for lower dimensional hidden layers than we used.  

Our results on these small and medium sized datasets confirm that overfitting is not observed with the DGP model, and that the DGP is never worse and often better than the single layer GP. We note in particular that on the `power', `protein' and `kin8nm' datasets all the DGP models outperform the SGP with five times the number of inducing points.


\paragraph{Rectangles Benchmark}
We use the Rectangle-Images dataset\footnote{\url{http://www.iro.umontreal.ca/~lisa/twiki/bin/view.cgi/Public/RectanglesData}}, which is specifically designed to distinguish deep and shallow architectures. The dataset consists of 12,000 training and 50,000 testing examples of size $28\times 28$, where each image consists of a (non-square) rectangular image against a different background image. The task is to determine which of the height and width is greatest. We run 2, 3 and 4 layer DGP models, and observe increasing performance with each layer. Table~\ref{table:rectangles} contains the results. Note that the 500 inducing point single-layer GP is significantly less effective than any of the deep models. Our 4-layer model achieves 77.9\% classification accuracy, exceeding the best result of 77.5\% reported in~\cite{Larochelle2007AnVariation} with a three-layer deep belief network. We also exceed the best result of 76.4\% reported in~\cite{Krauth2016AutoGP:Models} using a sparse GP with an Arcsine kernel, a leave-one-out objective, and 1000 inducing points. 

\begin{table}[h]
\centering
\caption{Results on Rectangles-Images dataset ($N = 12000$, $D=784$)}
\label{table:rectangles}
\scalebox{0.91}{
\begin{tabular}{lccccccccccccc}
\toprule
\multicolumn{1}{c}{} & 
\multicolumn{2}{c}{Single layer GP} & 
\multicolumn{3}{c}{Ours} &
\multicolumn{2}{c}{Larochelle [2007]} &
\multicolumn{1}{c}{Krauth [2016]} \\
\cmidrule(r){2-3} \cmidrule(r){4-6} \cmidrule(r){7-8} \cmidrule(r){9-9}  
& SGP & SGP 500 & DGP 2 & DGP 3 & DGP 4 & DBN-3 & SVM & SGP 1000\\
\midrule
Accuracy (\%) & $76.1$ & $76.4$ & $77.3$ & $77.8$ & $\textbf{77.9}$ & $77.5$ & $76.96$ & $76.4$ \\
Likelihood  & $-0.493$ & $-0.485$ & $0.475$ & $\mathbf{-0.460}$ & $\mathbf{-0.460}$ &-&-& $-0.478$\\
\bottomrule
\end{tabular}
}
\end{table}

\paragraph{Large-Scale Regression}
To demonstrate our method on a large scale regression problem we use the UCI `year' dataset and the `airline' dataset, which has been commonly used by the large-scale GP community. For the `airline' dataset we 
take the first 700K points for training and next 100K for testing. We use a random 10\% split for the `year' dataset. Results are shown in Table~\ref{table:regression_rmse}, with the log-likelihood reported in the supplementary material. In both datasets we see that the DGP models perform better with increased depth, significantly improving in both log likelihood and RMSE over the single-layer model, even with 500 inducing points.

\begin{table}[ht]
\centering
\caption{Regression test RMSE results for large datasets}
\label{table:regression_rmse}
\begin{tabular}{lllcccccc}
	\toprule
        & N      & D  & SGP     & SGP 500 & DGP 2   & DGP 3   & DGP 4   & DGP 5   \\
    \midrule
year    & 463810 & 90 & $10.67$ & $9.89$  & $9.58$  & $8.98$  & $8.93$  & $\mathbf{8.87}$   \\
airline & 700K   & 8  & $25.6$  & $25.1$  & $24.6$  & $24.3$  & $24.2$  & $\mathbf{24.1}$  \\
taxi    & 1B     & 9  & $337.5$ & $330.7$ & $281.4$ & $270.4$ & $268.0$ & $\mathbf{266.4}$ \\
    \bottomrule
\end{tabular}
\end{table}

\paragraph{MNIST Multiclass Classification}
We apply the DGP with 2 and 3 layers to the MNIST multiclass classification problem. We use the robust-max multiclass likelihood~\citep{Hernandez-Lobato2011RobustClassification} and use full unprocessed data with the standard training/test split of 60K/10K. The single-layer GP with 100 inducing points achieves a test accuracy of 97.48\% and this is increased to 98.06\% and 98.11\% with two and three layer DGPs, respectively. The 500 inducing point single layer model achieved 97.9\% in our implementation, though a slightly higher result for this model has previously been reported of 98.1\%~\citep{Hensman2013GaussianData} and 98.4\%~\citep{Krauth2016AutoGP:Models} for the same model with 1000 inducing points. We attribute this difference to different hyperparameter initialization and training schedules, and stress that we use exactly the same \initialization~and learning schedule for all our models. The only other DGP result in the literature on this dataset is 94.24\%~\citep{Wang2016SequentialProcess} for a two layer model with a two dimensional latent space.

\paragraph{Large-Scale Classification}
We use the HIGGS ($N=11$M, $D=28$) and SUSY ($N=5.5$M, $D=18$) datasets for large-scale binary classification. These datasets have been constructed from Monte Carlo physics simulations to detect the presence of the Higgs boson and super-symmetry~\citep{Baldi2014SearchingLearning}. We take a 10\% random sample for testing and use the rest for training. We use the AUC metric for comparison with~\cite{Baldi2014SearchingLearning}. Our DGP models are the highest performing on the SUSY dataset (AUC of 0.877 for all the DGP models) compared to shallow neural networks (NN, 0.875), deep neural networks (DNN, 0.876) and boosted decision trees (BDT, 0.863). On the HIGGS dataset we see a steady improvement in additional layers (0.830, 0.837, 0.841 and 0.846 for DGP 2--4 respectively). On this dataset the DGP models exceed the performance of BDT (0.810) and NN (0.816) and both single layer GP models SGP (0.785) and SGP 500 (0.794). The best performing model on this dataset is a 5 layer DNN (0.885). Full results are reported in the supplementary material. 

\begin{wraptable}{r}{4.0cm}
\vspace{-7mm}
\centering
\caption{Typical computation time in seconds for a single gradient step.} \label{profile_table}
\begin{tabular}{lccc}
\toprule
 & CPU & GPU  \\
\midrule
SGP & 0.14 & 0.018\\
SGP 500 & 1.71 & 0.11\\
DGP 2 & 0.36 &  0.030\\
DGP 3 & 0.49 & 0.045\\
DGP 4 & 0.65 &  0.056\\
DGP 5 & 0.87 & 0.069\\
\bottomrule
\end{tabular}
\vspace{-4mm}
\end{wraptable} 

\paragraph{Massive-Scale Regression}
To demonstrate the efficacy of our model on massive data we use the New York city yellow taxi trip dataset of 1.21 billion journeys \footnote{\url{http://www.nyc.gov/html/tlc/html/about/trip_record_data.shtml}}. Following~\cite{Peng2017AsynchronousProcesses} we use 9 features: time of day; day of the week; day of the month; month; pick-up latitude and longitude; drop-off latitude and longitude; travel distance. The target is to predict the journey time. We randomly select 1B ($10^9$) examples for training and use 1M examples for testing, and we scale both inputs and outputs to zero mean and unit standard deviation in the training data. We discard journeys that are less than 10\,s or greater than 5\,h, or start/end outside the New York region, which we estimate to have squared distance less than $5^o$ from the center of New York. The test RMSE results are the bottom row of Table~\ref{table:regression_rmse} and test log likelihoods are in the supplementary material. We note the significant jump in performance from the single layer models to the DGP.  
As with all the large-scale experiments, we see a consistent improvement extra layers, but on this dataset the improvement is particularly striking (DGP 5 achieves a 21\% reduction in RMSE compared to SGP)



\section{Related Work}
The first example of the outputs of a GP used as the inputs to another GP can be found in~\cite{Lawrence2007HierarchicalModels}. MAP approximation was used for inference. The seminal work of~\cite{Titsias2010BayesianModel} demonstrated how sparse variational inference could be used to propagate Gaussian inputs through a GP with a Gaussian likelihood. This approach was extended in~\cite{Damianou2011VariationalSystems} to perform approximate inference in the model of~\cite{Lawrence2007HierarchicalModels}, and shortly afterwards in a similar model~\cite{Lazaro-Gredilla2012BayesianProcesses}, which also included a linear mean function. The key idea of both these approaches is the \factorization~of the variational posterior \emph{between} layers. A more general model (flexible in depth and dimensions of hidden layers) introduced the term `DGP' and used a posterior that also \factorized~\emph{between} layers. These approaches require a linearly increasing number of variational parameters in the number of data. For high-dimensional observations, it is possible to \amortize~the cost of this \optimization~with an auxiliary model. This approach is pursued in~\cite{Dai2016VariationalProcesses}, and with a recurrent architecture in~\cite{Mattos2016RecurrentProcesses}. Another approach to inference in the exact model was presented in \cite{Hensman2014NestedProcesses}, where a sparse approximation was used within layers for the GP outputs, similar to \cite{Damianou2013DeepProcesses}, but with a projected distribution over the inputs to the next layer. The particular form of the variational distribution was chosen to admit a tractable bound, but imposes a constraint on the flexibility.

An alternative approach is to modify the DGP prior directly and perform inference in a parametric model. This is achieved in~\cite{Bui2016DeepPropagation} with an inducing point approximation within each layer, and in~\cite{Cutajar2016RandomProcesses} with an approximation to the spectral density of the kernel. Both approaches then apply additional approximations to achieve tractable inference. In~\cite{Bui2016DeepPropagation}, an approximation to expectation propagation is used, with additional Gaussian approximations to the log partition function to propagate uncertainly through the non-linear GP mapping. In~\cite{Cutajar2016RandomProcesses} a fully \factorized~ variational approximation is used for the spectral components. Both these approaches require specific kernels: in~\cite{Bui2016DeepPropagation} the kernel must have analytic expectations under a Gaussian, and in~\cite{Cutajar2016RandomProcesses} the kernel must have an analytic spectral density. \cite{Vafa2016TrainingSampling} also uses the same initial approximation as~\cite{Bui2016DeepPropagation} but applies MAP inference for the inducing points, such that the uncertainty propagated through the layers only represents the quality of the approximation. In the limit of infinitely many inducing points this approach recovers a deterministic radial basis function network. A particle method is used in~\cite{Wang2016SequentialProcess}, again employing an online version of the sparse approximation used by~\cite{Bui2016DeepPropagation} within each layer. Similarly to our approach, in ~\cite{Wang2016SequentialProcess} samples are taken through the conditional model, but differently from us they then use a point estimate for the latent variables. It is not clear how this approach propagates uncertainty through the layers, since the GPs at each layer have point-estimate inputs and outputs.  

A pathology with the DGP with zero mean function for the inner layers was identified in \cite{Duvenaud2014AvoidingNetworks}. In \cite{Duvenaud2014AvoidingNetworks} a suggestion was made to concatenate the original inputs at each layer. This approach is followed in \cite{Dai2016VariationalProcesses} and~\cite{Cutajar2016RandomProcesses}. 
The linear mean function was original used by~\cite{Lazaro-Gredilla2012BayesianProcesses}, though in the special case of a two layer DGP with a 1D hidden layer. To the best of our knowledge there has been no previous attempt to use a linear mean function for all inner layers.

\section{Discussion}
Our experiments show that on a wide range of tasks the DGP model with our doubly stochastic inference is both effective and scalable. Crucially, we observe that on the small datasets the DGP does not overfit, while on the large datasets additional layers generally increase performance and never deteriorate it. In particular, we note that the largest gain with increasing layers is achieved on the largest dataset (the taxi dataset, with 1B points). We note also that on all the large scale experiments the SGP 500 model is outperformed by the \emph{all} the DGP models. Therefore, for the same computational budget increasing the number of layers can be significantly more effective than increasing the accuracy of approximate inference in the single-layer model. Other than the additional computation time, which is fairly modest (see Table~\ref{profile_table}), we do not see downsides to using a DGP over a single-layer GP, but substantial advantages. 

While we have considered simple kernels and black-box applications, any domain-specific kernel could be used in any layer. This is in contrast to other methods \citep{Damianou2013DeepProcesses,Bui2016DeepPropagation,Cutajar2016RandomProcesses} that require specific kernels and intricate implementations. Our implementation is simple (< 200 lines), publicly available~\footnote{\url{https://github.com/ICL-SML/Doubly-Stochastic-DGP}}, and is integrated with GPflow~\citep{2017GPflow}, an open-source GP framework built on top of Tensorflow~\citep{Abadi2015TensorFlow:Systems}.

\section{Conclusion}
We have presented a new method for inference in Deep Gaussian Process (DGP) models. With our inference we have shown that the DGP can be used on a range of regression and classification tasks with no hand-tuning. Our results show that in practice the DGP always exceeds or matches the performance of a single layer GP. Further, we have shown that the DGP often exceeds the single layer significantly, even when the quality of the approximation to the single layer is improved. Our approach is highly scalable and benefits from GPU acceleration. 

The most significant limitation of our approach is the dealing with high dimensional inner layers. We used a linear mean function for the high dimensional datasets but left this mean function fixed, as to \optimize~the parameters would go against our non-parametric paradigm. It would be possible to treat this mapping probabilistically, following the work of~\citet{Titsias2013VariationalRegression}.

\subsection*{Acknowledgments}
We have greatly appreciated valuable discussions with James Hensman and Steindor Saemundsson in the preparation of this work. We thank Vincent Dutordoir and anonymous reviewers for helpful feedback on the manuscript. We are grateful for a Microsoft Azure Scholarship and support through a Google Faculty Research Award to Marc Deisenroth.

\medskip
\bibliographystyle{abbrvnat}
\bibliography{bib}

\newpage
\section*{Supplementary Material}

\section*{Experiment Details}
\label{Appendix_A}
\textbf{Variational parameters initializations}
All our DGP models have 100 inducing points, initialized with K-means (computed from a random 100M subset for the taxi data). The inducing function values means are all initialized to zero, and variances to the identity, which we scale by $10^{-5}$ for the inner layers. 

\textbf{Model hyperparameter initializations}
We initialize all kernel variances and lengthscales to 2 at every layer, and initialize the likelihood variance to 0.01. We initialize the noise between the layers (separately for each layer) to $10^{-5}$.

\textbf{Training}
We optimize all hyperparameters and variational parameters jointly. We use a minibatch size of 10K (or the size of the data for the datasets with fewer than 10K points). We use the Adam optimizer~\citep{Kingma2014Adam:Optimization} with a learning rate of 0.01 with all other parameters set to the default values. We train for 20,000 iterations for the small to medium regression datasets, and 100,000 for the others (500,000 for the taxi dataset, which is 5 epochs)

\begin{wraptable}{r}{5.0cm}
\vspace{-7mm}
\centering
\caption{Typical computation time in seconds for a single gradient step} \label{profile_table_appendix}
\begin{tabular}{lccc}
\toprule
 & CPU & GPU  \\
\midrule
SGP & 0.14 & 0.018\\
SGP 500 & 1.71 & 0.11\\
DGP 2 & 0.36 &  0.030\\
DGP 3 & 0.49 & 0.045\\
DGP 4 & 0.65 &  0.056\\
DGP 5 & 0.87 & 0.069\\
\bottomrule
\end{tabular}
\vspace{-4mm}
\end{wraptable} 

\textbf{Computation}
Our implementation is based on GPflow~\cite{2017GPflow} and leverages automatic differentiation and GPU acceleration in Tensorflow~\citep{Abadi2015TensorFlow:Systems}. We use Azure NC6 instances with Tesla K80 GPUs for all computations. The GPU implementation speeds up computation by an order of magnitude. See Table~\ref{profile_table_appendix} (repeated here from the main text) for timing results for CPU (8 core i5) and GPU (Tesla K80). These timing results are for a minibatch size of 10000, with inner dimensions all equal to one, averaged over 100 steps. Note that we achieve slightly sub-linear scaling in depth.  

\clearpage
\section*{Further results}

\begin{figure}[h]
\includegraphics[width=\textwidth]{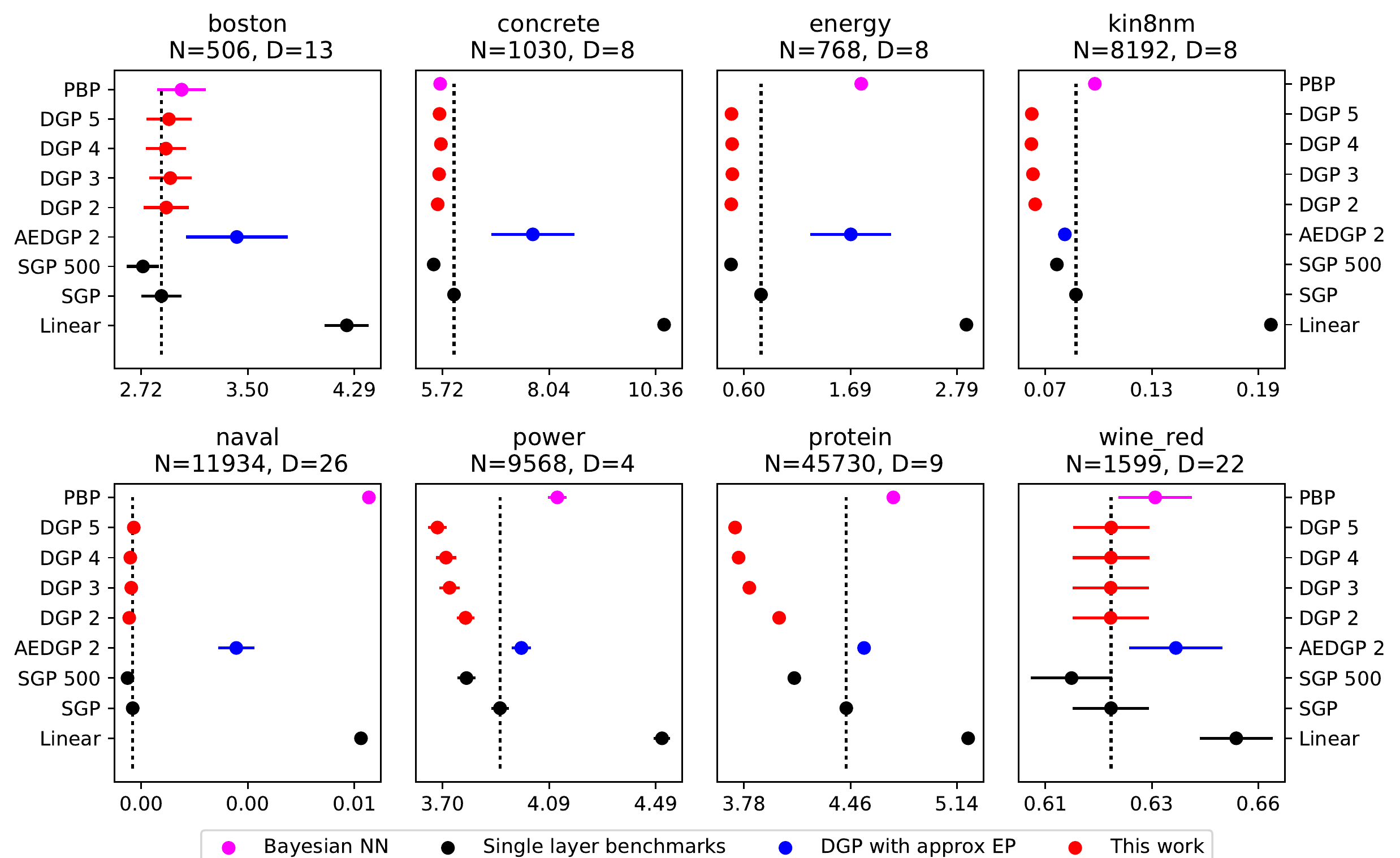}
\caption{Regression test RMSE results on benchmark datasets. Lower (to the left) is better. The mean is shown with error bars of one standard error. The sparse GP with the same number of inducing points is highlighted as a baseline.
}\label{fig:reg_results_rmse}
\end{figure}

\vspace{1cm}

\begin{table}[h]
\centering
\caption{Regression test log likelihood results for large datasets}
\label{table:regression_lik_appendix}
\scalebox{1.0}{
\begin{tabular}{lllcccccc}
	\toprule
        & N      & D  & SGP     & SGP 500 & DGP 2   & DGP 3   & DGP 4   & DGP 5   \\
    \midrule
year    & 463810 & 90 & $-3.74$ & $-3.65$ & $-3.63$ & $-3.57$ & $-3.56$ &       $\mathbf{-3.32}$  \\
airline & 700K   & 8  & $-4.66$ & $-4.63$ & $-4.61$ & $-4.59$ & $-4.59$ & $\mathbf{-4.58}$  \\
taxi    & 1B     & 9  & $-7.24$ & $-7.22$ & $-7.06$ & $-7.02$ & $-7.01$ & $\mathbf{-7.00}$ \\
    \bottomrule
\end{tabular}
}
\end{table}

\vspace{1cm} 


\begin{table}[h]
\caption{Binary classification AUC results on high energy physics data. We report vales for comparison with~\cite{Baldi2014SearchingLearning}} \label{table:binary_classification}
\begin{center}
\scalebox{0.92}{
\begin{tabular}{lccccccccccccccc}
\toprule
\multicolumn{3}{c}{} & 
\multicolumn{2}{c}{Single layer GP} & 
\multicolumn{4}{c}{Ours} &
\multicolumn{3}{c}{Other reported results} \\
\cmidrule(r){4-5} \cmidrule(r){6-9} \cmidrule(r){10-12} 
& N & D & SGP & SGP 500 & DGP 2 & DGP 3 & DGP 4 & DGP 5 & BDT & NN & DNN \\
\midrule
HIGGS & $11\text{M}$ &  24 & 
0.785 & 
0.794 & 
0.830 & 
0.837 & 
0.841 & 
0.846 & 
0.810  & 
0.816 & 
\textbf{0.885} \\ 
SUSY &
$5.5\text{M}$ &
18 & 
0.875 & 
0.876 & 
\textbf{0.877} & 
\textbf{0.877} & 
\textbf{0.877} & 
\textbf{0.877} & 
0.863  & 
0.875 & 
0.876 \\ 
\bottomrule
\end{tabular}
}
\end{center}
\end{table}

\begin{landscape}

\begin{table}[h]
\centering
\caption{Regression test log likelihood results. Reported is the mean over 20 splits (with standard errors)}\label{table:reg_lik_appendix}
\scalebox{0.9}{
\begin{tabular}{lllccccccccc}
\toprule
          & N     & D  & Linear        & SGP           & SGP 500       & AEPDGP 2      & DGP 2         & DGP 3         & DGP 4         & DGP 5         & PBP           \\
 \midrule
boston    & 506   & 13 & $-2.89(0.03)$ & $\mathbf{-2.47(0.05)}$ & $\mathbf{-2.40(0.07)}$ & $\mathbf{-2.46(0.09)}$ & $\mathbf{-2.47(0.05)}$ & $-2.49(0.05)$ & $-2.48(0.05)$ & $-2.49(0.05)$ & $\mathbf{-2.57(0.09)}$ \\
concrete  & 1030  & 8  & $-3.78(0.01)$ & $-3.18(0.02)$ & $\mathbf{-3.09(0.02)}$ & $-3.30(0.12)$ & $-3.12(0.01)$ & $-3.13(0.01)$ & $-3.14(0.01)$ & $-3.13(0.01)$ & $-3.16(0.02)$ \\
energy    & 768   & 8  & $-2.48(0.02)$ & $-1.29(0.02)$ & $\mathbf{-0.63(0.03)}$ & $-1.65(0.15)$ & $-0.73(0.02)$ & $-0.75(0.02)$ & $-0.76(0.02)$ & $-0.74(0.02)$ & $-2.04(0.02)$ \\
kin8nm    & 8192  & 8  & $0.18(0.01)$  & $0.97(0.00)$  & $1.15(0.00)$  & $1.15(0.03)$  & $1.34(0.01)$  & $1.37(0.01)$  & $\mathbf{1.38(0.01)}$  & $\mathbf{1.38(0.01)}$  & $0.90(0.01)$  \\
naval     & 11934 & 26 & $3.73(0.00)$  & $6.57(0.15)$  & $\mathbf{7.01(0.05)}$  & $4.37(0.23)$  & $6.76(0.19)$  & $6.62(0.18)$  & $6.61(0.17)$  & $6.41(0.28)$  & $3.73(0.01)$  \\
power     & 9568  & 4  & $-2.93(0.01)$ & $-2.79(0.01)$ & $-2.75(0.01)$ & $-2.78(0.01)$ & $-2.75(0.01)$ & $\mathbf{-2.74(0.01)}$ & $\mathbf{-2.74(0.01)}$ & $\mathbf{-2.73(0.01)}$ & $-2.84(0.01)$ \\
protein   & 45730 & 9  & $-3.07(0.00)$ & $-2.91(0.00)$ & $-2.83(0.00)$ & $-2.81(0.01)$ & $-2.81(0.00)$ & $-2.75(0.00)$ & $-2.73(0.00)$ & $\mathbf{-2.71(0.00)}$ & $-2.97(0.00)$ \\
wine\_red & 1599  & 22 & $-0.99(0.01)$ & $-0.95(0.01)$ & $\mathbf{-0.93(0.01)}$ & $-1.51(0.09)$ & $-0.95(0.01)$ & $-0.95(0.01)$ & $-0.95(0.01)$ & $-0.95(0.01)$ & $-0.97(0.01)$ \\
\bottomrule
\end{tabular}
}
\end{table}

\begin{table}[h]
\centering
\caption{Regression test RMSE results}\label{table:reg_lik}
\scalebox{0.9}{
\begin{tabular}{lllccccccccc}
\toprule
& N     & D  & Linear        & SGP          & SGP 500      & AEPDGP 2     & DGP 2        & DGP 3        & DGP 4        & DGP 5        & PBP          \\
\midrule
boston    & 506   & 13 & $4.24(0.16)$  & $2.87(0.15)$ & $\mathbf{2.73(0.12)}$ & $3.42(0.37)$ & $2.90(0.17)$ & $2.93(0.16)$ & $2.90(0.15)$ & $2.92(0.17)$ & $3.01(0.18)$ \\
concrete  & 1030  & 8  & $10.54(0.13)$ & $5.97(0.11)$ & $\mathbf{5.53(0.12)}$ & $7.68(0.90)$ & $\mathbf{5.61(0.10)}$ & $\mathbf{5.64(0.10)}$ & $5.68(0.10)$ & $5.65(0.10)$ & $5.67(0.09)$ \\
energy    & 768   & 8  & $2.88(0.05)$  & $0.78(0.02)$ & $\mathbf{0.47(0.02)}$ & $1.70(0.42)$ & $\mathbf{0.47(0.01)}$ & $\mathbf{0.48(0.01)}$ & $\mathbf{0.48(0.01)}$ & $\mathbf{0.47(0.01)}$ & $1.80(0.05)$ \\
kin8nm    & 8192  & 8  & $0.20(0.00)$  & $0.09(0.00)$ & $0.08(0.00)$ & $0.08(0.00)$ & $\mathbf{0.06(0.00)}$ & $\mathbf{0.06(0.00)}$ & $\mathbf{0.06(0.00)}$ & $\mathbf{0.06(0.00)}$ & $0.10(0.00)$ \\
naval     & 11934 & 26 & $0.01(0.00)$  & $\mathbf{0.00(0.00)}$ & $\mathbf{0.00(0.00)}$ & $\mathbf{0.00(0.00)}$ & $\mathbf{0.00(0.00)}$ & $\mathbf{0.00(0.00)}$ & $\mathbf{0.00(0.00)}$ & $\mathbf{0.00(0.00)}$ & $0.01(0.00)$ \\
power     & 9568  & 4  & $4.51(0.03)$  & $3.91(0.03)$ & $3.79(0.03)$ & $3.99(0.03)$ & $3.79(0.03)$ & $3.73(0.04)$ & $\mathbf{3.71(0.04)}$ & $\mathbf{3.68(0.03)}$ & $4.12(0.03)$ \\
protein   & 45730 & 9  & $5.21(0.02)$  & $4.43(0.03)$ & $4.10(0.03)$ & $4.54(0.02)$ & $4.00(0.03)$ & $3.81(0.04)$ & $3.74(0.04)$ & $\mathbf{3.72(0.04)}$ & $4.73(0.01)$ \\
wine\_red & 1599  & 22 & $0.65(0.01)$  & $0.63(0.01)$ & $\mathbf{0.62(0.01)}$ & $0.64(0.01)$ & $0.63(0.01)$ & $0.63(0.01)$ & $0.63(0.01)$ & $0.63(0.01)$ & $0.64(0.01)$ \\
\bottomrule
\end{tabular}
}
\end{table}

\end{landscape}

\section*{Derivation of the Lower Bound}
The evidence lower bound of our DGP model is given by
\begin{align*}
&\mathcal L_{DGP} = \mathbb{E}_{\red{q(\{\layer{\vec F}{l}, \layer{\vec U}{l}\}_{l=1}^L)}}\left[\frac{\blue{p(\vec Y,\{\layer{\vec F}{l}, \layer{\vec U}{l}\}_{l=1}^L)}}{\red{q(\{\layer{\vec F}{l}, \layer{\vec U}{l}\}_{l=1}^L)}}\right]\,,\\
&\red{q(\{\layer{\vec F}{l}, \layer{\vec U}{l}\}_{l=1}^L)} = \prod_{l=1}^L p(\layer{\vec F}{l}|\layer{\vec U}{l}; \layer{\vec F}{l-1}; \layer{\vec Z}{l-1})q(\layer{\vec U}{l})\\
& \blue{p(\vec Y,\{\layer{\vec F}{l}, \layer{\vec U}{l}\}_{l=1}^L)} = \underbrace{
\prod\nolimits_{i=1}^{N}p(\vec y_i|\layer{\vec f_i}{L})
}_{\text{likelihood}}
\underbrace{
\prod\nolimits_{l=1}^{L} 
p( \layer{\vec{F}}{l}|  \layer{\vec {U}}{l};  \layer{\vec{F}}{l-1},  \layer{\vec{Z}}{l-1})
p( \layer{\vec{U}}{l} ;  \layer{\vec{Z}}{l-1})
}_{\text{DGP prior}}
\end{align*}
Therefore,
\begin{align*}
\mathcal L_{DGP}  & = \iint \red{q(\{\layer{\vec F}{l}, \layer{\vec U}{l}\}_{l=1}^L)} \log\left(
\frac{
\blue{p(\vec Y,\{\layer{\vec F}{l}, \layer{\vec U}{l}\}_{l=1}^L)}
}
{
\red{q(\{\layer{\vec F}{l}, \layer{\vec U}{l}\}_{l=1}^L)}
}
\right)
d\{\layer{\vec F}{l}, \layer{\vec U}{l}\}_{l=1}^L \\
&=\iint \red{q(\{\layer{\vec F}{l}, \layer{\vec U}{l}\}_{l=1}^L)} \\
& \log\left(
\frac{
\blue{\prod\nolimits_{i=1}^{N}p(\vec y_i|\layer{\vec f_i}{L})\prod\nolimits_{l=1}^{L} 
p( \layer{\vec{F}}{l}|  \layer{\vec {U}}{l};  \layer{\vec{F}}{l-1},  \layer{\vec{Z}}{l-1})
p( \layer{\vec{U}}{l} ;  \layer{\vec{Z}}{l-1})}
}
{
\red{\prod_{l=1}^L p(\layer{\vec F}{l}|\layer{\vec U}{l}; \layer{\vec F}{l-1}; \layer{\vec Z}{l-1})q(\layer{\vec U}{l})}
}
\right)
d\{\layer{\vec F}{l}, \layer{\vec U}{l}\}_{l=1}^L
\end{align*}
We see that terms inside the logarithm cancel out, such that we obtain
\begin{align*}
\mathcal L_{DGP}&=\iint \red{q(\{\layer{\vec F}{l}, \layer{\vec U}{l}\}_{l=1}^L)} \log\left(
\frac{
\blue{\prod\nolimits_{i=1}^{N}p(\vec y_i|\layer{\vec f_i}{L})\prod\nolimits_{l=1}^{L} 
p( \layer{\vec{U}}{l} ;  \layer{\vec{Z}}{l-1})}
}
{
\red{\prod_{l=1}^L q(\layer{\vec U}{l})}
}
\right)
d\{\layer{\vec F}{l}, \layer{\vec U}{l}\}_{l=1}^L\\
&=\orange{\iint q(\{\layer{\vec F}{l}, \layer{\vec U}{l}\}_{l=1}^L) \log\left(
\prod\nolimits_{i=1}^{N}p(\vec y_i|\layer{\vec f_i}{L})
\right)
d\{\layer{\vec F}{l}, \layer{\vec U}{l}\}_{l=1}^L
}\\
&\quad +\green{\iint q(\{\layer{\vec F}{l}, \layer{\vec U}{l}\}_{l=1}^L) \log\left(
\frac{
\prod\nolimits_{l=1}^{L} 
p( \layer{\vec{U}}{l} ;  \layer{\vec{Z}}{l-1})
}
{
\prod_{l=1}^L q(\layer{\vec U}{l})
}
\right)
d\{\layer{\vec F}{l}, \layer{\vec U}{l}\}_{l=1}^L
}\\
&=\orange{
\int q(\{\layer{\vec F}{l}\}_{l=1}^L) \log\left(
\prod\nolimits_{i=1}^{N}p(\vec y_i|\layer{\vec f_i}{L})
\right)
d\{\layer{\vec F}{l}\}_{l=1}^L
}
\\
&\quad +
\green{
\int q(\{\layer{\vec U}{l}\}_{l=1}^L) \log\left(
\frac{
\prod\nolimits_{l=1}^{L} 
p( \layer{\vec{U}}{l} ;  \layer{\vec{Z}}{l-1})
}
{
\prod_{l=1}^L q(\layer{\vec U}{l})
}
\right)
d\{\layer{\vec U}{l}\}_{l=1}^L
}\\
&=
\orange{
\mathbb{E}_{q(\layer{\vec f_i}{L})}[\log p(y_i|\layer{\vec f_i}{L})] 
}
-
\green{
\sum\nolimits_{l=1}^L KL\big(q(\layer{\vec U}{l})||p( \layer{\vec{U}}{l} ;  \layer{\vec{Z}}{l-1})\big)
}
\end{align*}

\end{document}